\documentclass[pdflatex,sn-basic]{sn-jnl}
\usepackage{graphicx}%
\usepackage{multirow}%
\usepackage{amsmath,amssymb,amsfonts, bm}%
\usepackage{amsthm}%
\usepackage{mathrsfs}%
\usepackage[title]{appendix}%
\usepackage{xcolor}%
\usepackage{textcomp}%
\usepackage{manyfoot}%
\usepackage{booktabs}%
\usepackage{algorithm}%
\usepackage{algorithmicx}%
\usepackage{algpseudocode}%
\usepackage{listings}%
\usepackage{subfigure}  
\usepackage{enumitem}
\usepackage{url}

\begin{document}

\title[Article Title]{A method for classification of data with uncertainty using hypothesis testing}

\author[1]{\fnm{Shoma} \sur{Yokura}}\email{sd230006@cis.fukuoka-u.ac.jp}
\author*[1]{\fnm{Akihisa} \sur{Ichiki}}\email{ichiki@fukuoka-u.ac.jp}
\equalcont{These authors contributed equally to this work.}

\affil{\orgdiv{Department of Applied Mathematics}, \orgname{Faculty of Science}, \orgaddress{\street{Fukuoka University, 8-19-1, Nanakuma, Jonan-ku}, \city{Fukuoka City}, \postcode{814-0180}, \state{Fukuoka Prefecture}, \country{Japan}}}

\abstract{Binary classification is widely utilized in various fields to classify data into one of two distinct classes. However, conventional classifiers tend to make overconfident predictions for data that belong to overlapping regions of the two class distributions or for data outside the distributions (out-of-distribution data). Therefore,
conventional classifiers should not be applied in high-risk fields where classification results can have significant consequences.
In order to address this issue, it is necessary to quantify uncertainty and adopt decision-making approaches that take it into
account. Many methods have been proposed for this purpose; however, implementing these methods often requires performing
resampling, modifying the structure or performance of models, and optimizing the thresholds of classifiers. We propose a
new decision-making approach using two types of hypothesis testing. This method is capable of detecting ambiguous data
that belong to the overlapping regions of two class distributions, as well as out-of-distribution data that are not included in
the training data distribution. In addition, we quantify uncertainty using the empirical distribution of feature values derived
from the training data obtained through the trained model.  The classification threshold is determined by the $\alpha$ quantile and
(1$-\alpha$) quantile, where the significance level $\alpha$ is set according to each specific situation.}




\maketitle
\section{Introduction}\label{sec1}
Binary classification is a task to classify data into two different classes and has been applied in various fields. 
However, conventional classifiers tend to assign data to one of the classes, even for data points that fall within overlapping regions of the two class distributions or lie outside the distribution. In such cases, the classifier may make overconfident predictions, leading to an increased risk of misclassification, particularly for uncertain data. In high-risk domains, there are situations where conventional classifiers should be avoided. For example, in the medical field, ambiguous cases often arise during image diagnosis. In such scenarios, traditional classifiers lack the ability to respond with “I don’t know" when faced with uncertain data \citep{1,2}. To address this issue, many methods have been proposed for quantifying uncertainty and improving the decision-making process.\\
\indent In the field of uncertainty quantification, Bayesian inference methods are well-known. While applying Bayesian Neural Networks (BNN) \citep{3,4} tends to increase computational costs, recent methods have been proposed to reduce it by applying Monte Carlo Dropout (MC-Dropout) \citep{6,7}, which is based on the Dropout \citep{5} technique originally proposed for regularization. This approach enables uncertainty quantification while keeping computational costs lower. In the decision-making process, a method called Classification with reject option \citep{8,9} has been proposed. In this approach, if the value of the confidence function is below a pre-set threshold, the prediction is rejected. The confidence function is often defined based on the variance from MC-Dropout \citep{10,11} or the softmax response \citep{12}. Additionally, the threshold is set as a parameter and requires optimization.\\
\indent We propose a novel decision-making approach using two types of hypothesis testing. This method employs features obtained from a model trained on the training data. When applying this method to a binary classification problem, the distribution of each class can be approximated by the empirical distribution of the features obtained from the training data, and uncertainty can be quantified based on this empirical distribution. The threshold for classification is determined by the $\alpha$ quantile and the $1-\alpha$ quantile, which are based on the significance level $\alpha$ of the hypothesis testing. Our method has the ability to detect data belonging to the overlapping region of the two class distributions as well as out-of-distribution data. Furthermore, in uncertainty quantification, there is no need for resampling or model modefication. By using hypothesis testing, the threshold is determined based on the significance level. Since this significance level is appropriately set according to the context, there is no need for optimization, which helps to reduce computational costs.\\
\indent The structure of this paper is as follows: In the next section we introduce the two types of hypothesis testing. We define the test statistics and their distributions. In Section \ref{sec3}, we apply our method to the binary classification task of spiral pattern data. We examine whether our method captures the uncertainty. In Section \ref{sec4}, we apply our method to the task of binary classification of chest X-ray images as negative or positive for pneumonia. In the last section, we discuss the experimental results of Sections \ref{sec3} and \ref{sec4} and conclude. 

\section{Proposed method}\label{sec2}
\subsection{Two types of hypothesis testing}\label{subsec1}
Let $x$ be a vector in $\mathbb{R}^D$ and $g : \mathbb{R}^D\to \mathbb{R}$ be a map. Let $C_1$ and $C_2$ be the classes 
in the binary classification problem, and let $F_1(z)$ and $F_2(z)$ ($z\in \mathbb{R}$) be probability 
distributions that the feature values of the respective classes follow. For $i = 1, 2$, let $H^i_0$ be 
the null hypothesis that the data $x$ belongs to $C_i$ and $H^i_1$ be the alternative 
hypothesis that $x$ does not belong to $C_i$. The following two types of hypothesis testing are 
conducted on the feature $g(x)$:
\begin{align}
    H_0^1:\quad g(x) \sim F_1, \quad H_1^1: \quad g(x)\not\sim F_1, \label{eq1}\\
    H_0^2:\quad g(x) \sim F_2, \quad H_1^2:\quad g(x)\not\sim F_2.\label{eq2}
\end{align}
The hypothesis testing \eqref{eq1} verifies whether the data $x$ belongs to class $C_1$, and the 
hypothesis testing \eqref{eq2} does whether it belongs to class $C_2$. The rejection regions of the aforementioned 
hypothesis testing methods are denoted by $I_1$ and $I_2$, respectively, and the acceptance regions are defined as 
$I^c_1 = \mathbb{R}\smallsetminus I_1$ and $I^c_2 = \mathbb{R}\smallsetminus I_2$. The two classes are deemed 
uncertain if both types of hypothesis testing can be rejected, i.e., 
if the value of the test statistic is an element of $I_1$ and $I_2$. Additionally, 
if neither type of hypothesis testing can be rejected, i.e., if the value of the test statistic 
is an element of $I^c_1$ and $I^c_2$, then the two classes are considered uncertain as well.

\subsection{Test statistic}\label{subsec2}
In the context of our hypothesis testing, the test statistic, denoted by $t$, 
is specified as follows: 
\begin{align}
    t=g(x),
\end{align}
where the map $g$ corresponds to the discriminant function in the case of Support Vector Machine (SVM) \citep{13,14}. In the context 
of binary classification with a neural network, the map $g$ that satisfies 
$\sigma\left(g(x)\right) = 1/(1 + \exp(-g(x)))$ is selected as the test statistic. The quantity 
$\sigma\left(g(x)\right)$ corresponds to the probability that $x$ belongs to the class $C_1$.

\subsection{Probability distribution of test statistic}\label{subsec3}
In the context of hypothesis testing, it is imperative to ascertain the probability distributions 
$F_1$ and $F_2$ for the two classes $C_1$ and $C_2$, respectively. However, the true probability 
distributions are not known. Assuming the utilization of supervised learning, class labels are assigned 
to the training data. Consequently, it becomes feasible to approximate the two probability distributions 
with the empirical ones.\\
\indent Let $D= \left\{(t_i, y_i) | t_i\in\mathbb{R}, y_i\in\{0, 1\}, i = 1, 2, ... , N\right\}$ 
be the set of training data, where $y_i$ denotes the label of the $i$-th data. The label $y_i=0$ corresponds to $C_1$, and $y_1=1$ does to $C_2$. Defining $D_1 = \{(t_n, 0)|n = 1, 2, ..., N_1\}\subset \mathcal{D}$ as the set consisting of 
data only in the class $C_1$, where $N_1$ is the number of samples in $C_1$. The distribution $F_1(t)$ of $C_1$ is approximated as follows: 
\begin{align}
    F_1(t) &\approx \hat{F}_1(t) \nonumber \\
    &=\frac{1}{N_1}\sum_{n=1}^{N_1}\delta(t-t_n),\label{eq3}
\end{align}
where $\delta(\cdot)$ is the Dirac delta function. The empirical distribution of $C_2$ is also approximated 
as $F_2 \approx \hat{F}_2$ in the same manner as in equation \eqref{eq3}. By approximating the true distributions with 
the corresponding empirical ones, it becomes possible to visualize the distributions using the histograms of 
the feature values. Visualization is an important key to setting the appropriate significance level for 
hypothesis testing. 
\section{Benchmark test}\label{sec3}
\begin{figure}[t]
    \centering
    \includegraphics[width=0.8\textwidth]{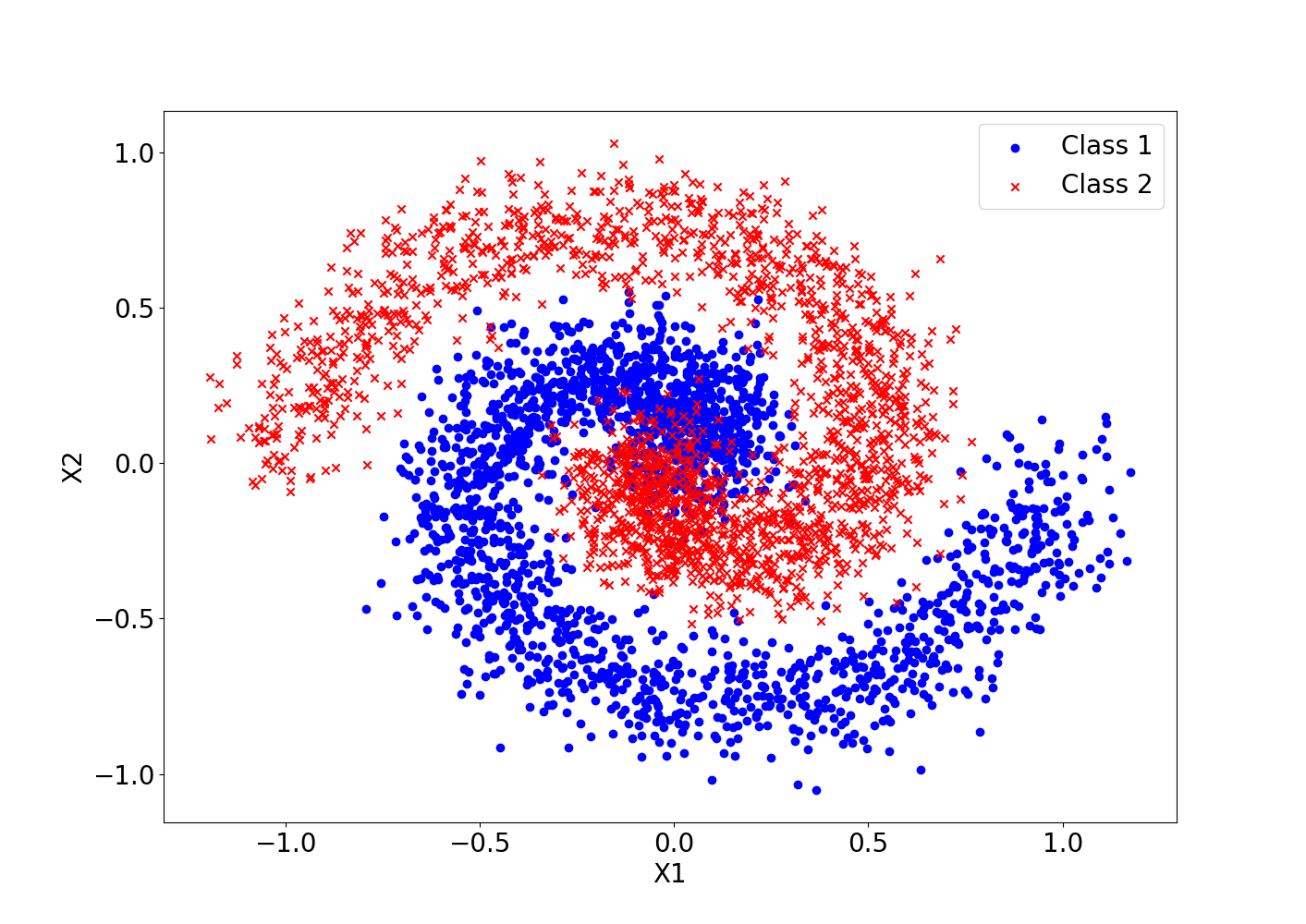}
    \caption{A dataset of spiral patterns. Dots represent $C_1$, while croses represent $C_2$ (Color figure online)}
    \label{fig:Fig1}
\end{figure}
\begin{figure}[t]
    \centering
    \subfigure[Histogram of test statistic]{
        \includegraphics[width=0.45\textwidth]{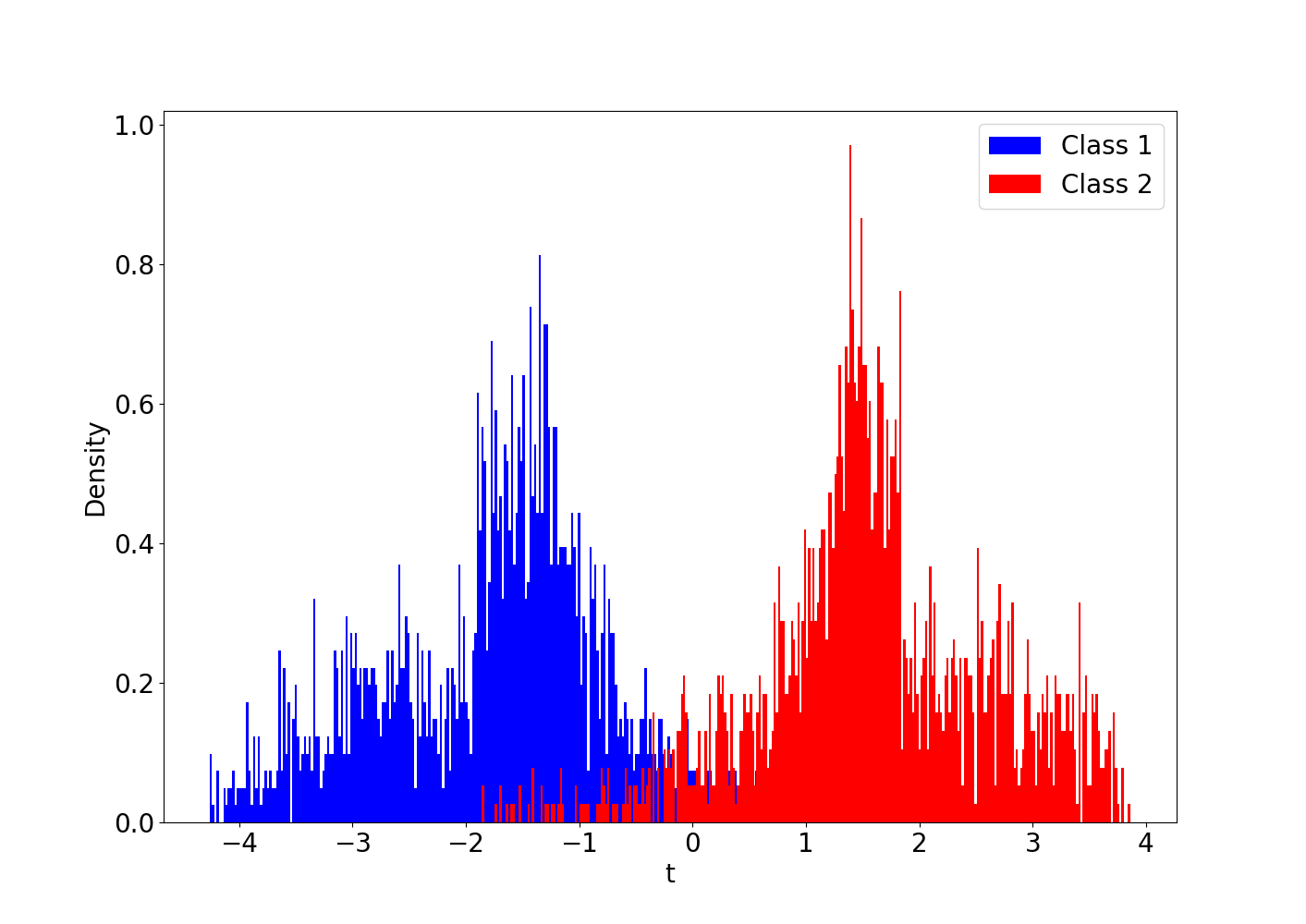}
        \label{fig:D1}
    }\hfill
    \subfigure[Acceptance regions of experiment \ref{item:a}]{
        \includegraphics[width=0.45\textwidth]{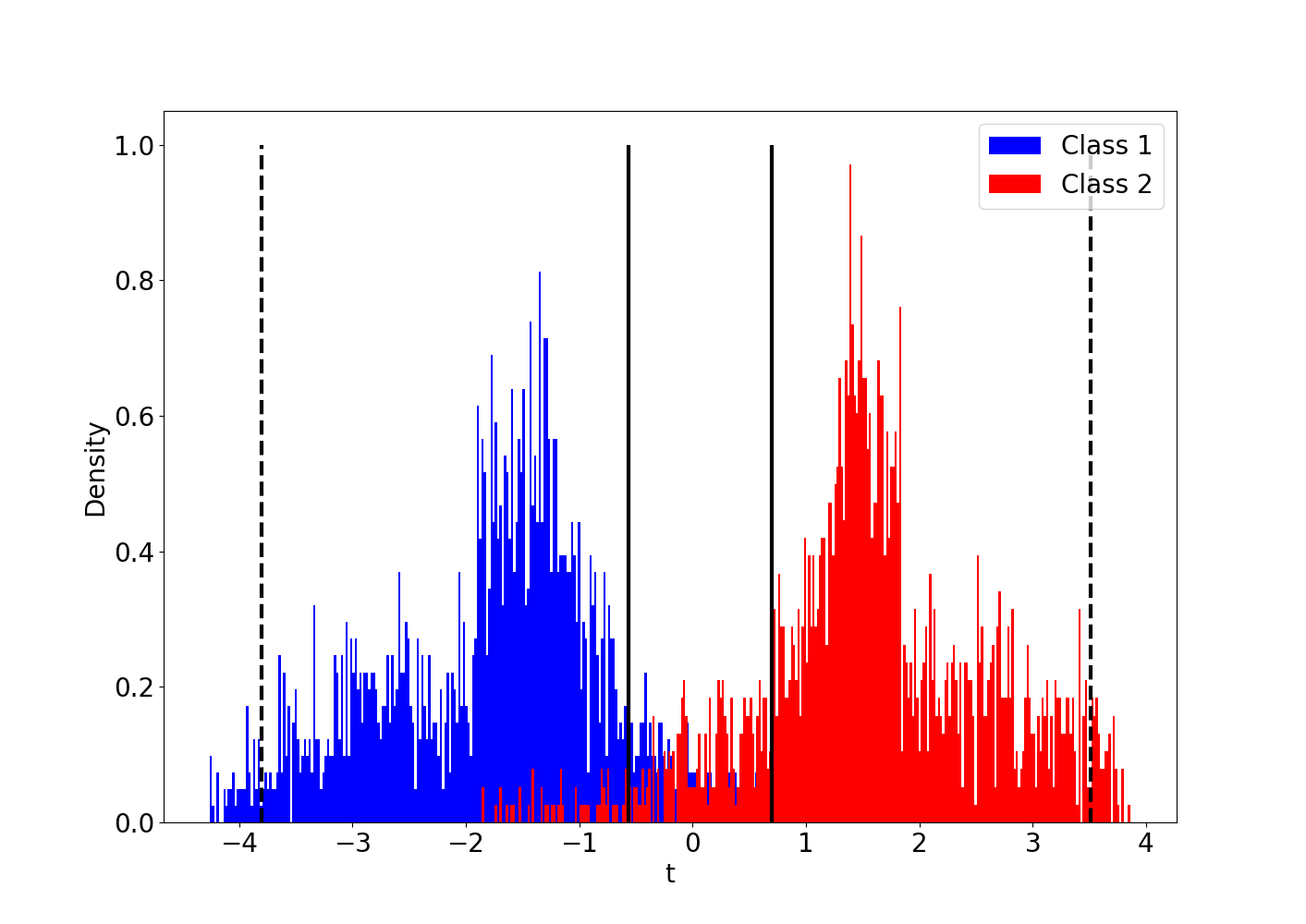}
        \label{fig:D2}
    } \\
    \subfigure[Acceptance regions of experiment \ref{item:b}]{
        \includegraphics[width=0.45\textwidth]{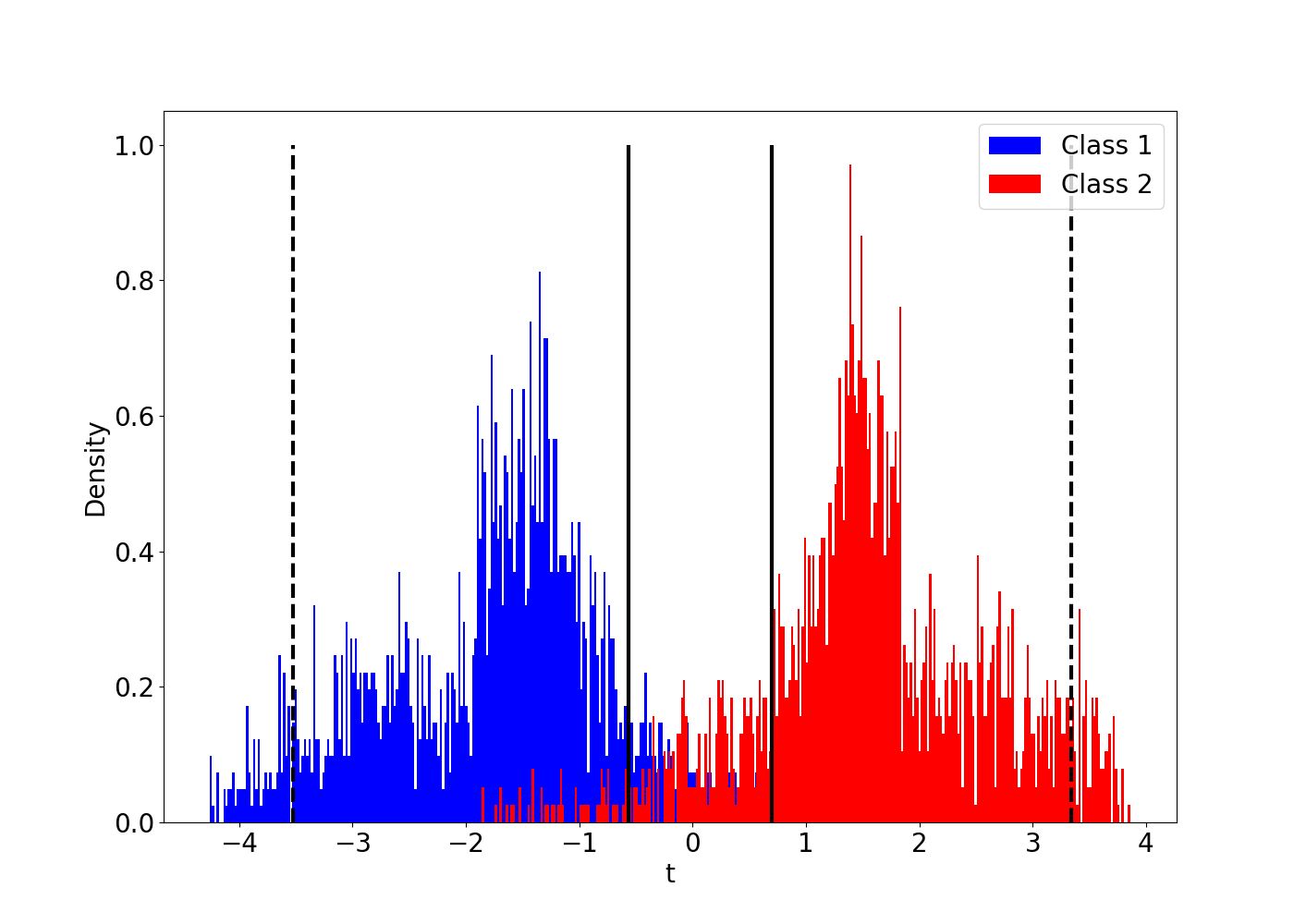}
        \label{fig:D3}
    }\hfill
    \subfigure[acceptance regions of experiment \ref{item:c}]{
        \includegraphics[width=0.45\textwidth]{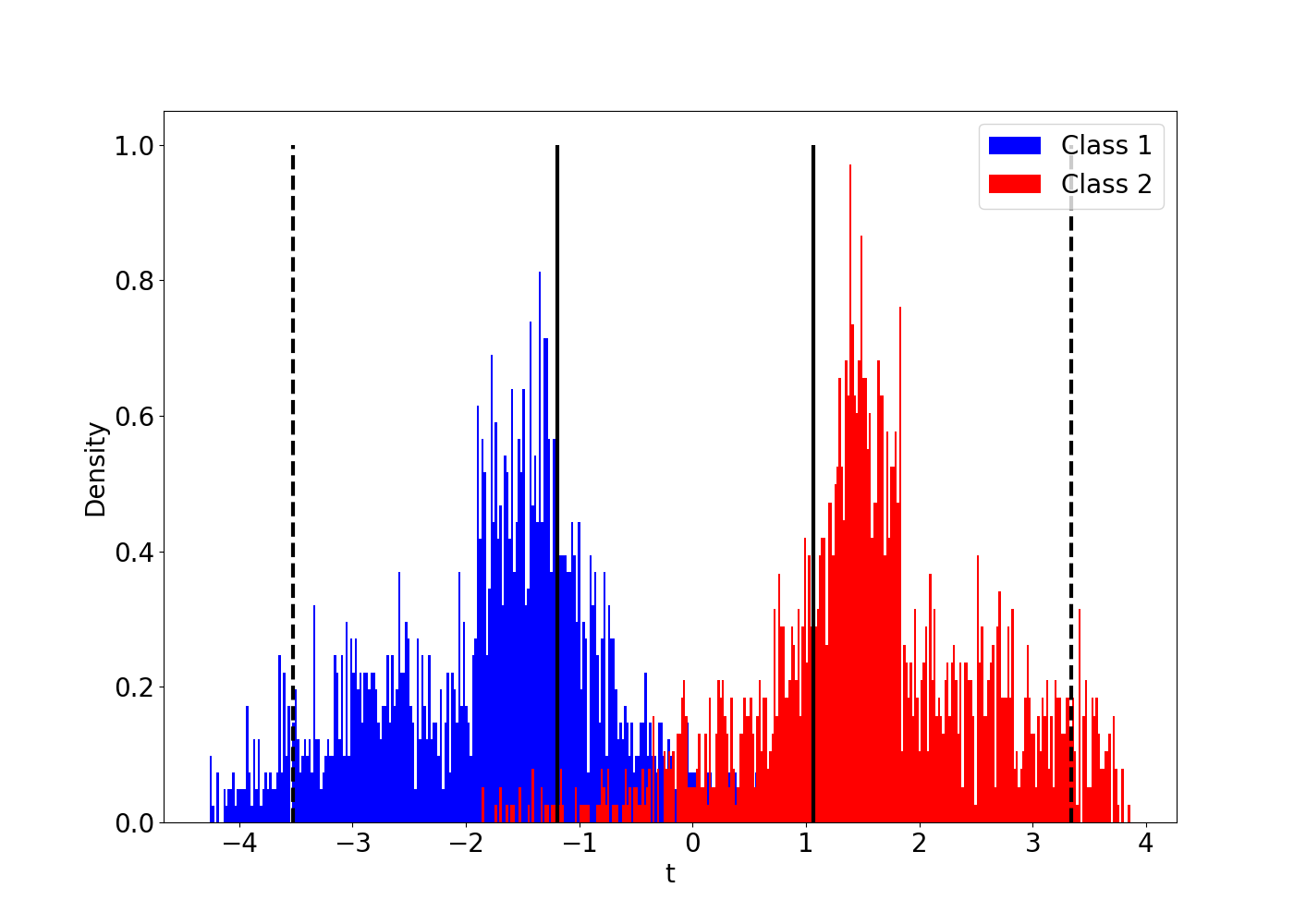}
        \label{fig:D4}
    }
    \caption{Histograms of the test statistics obtained from the training 
    data and visualization of the acceptance regions for each experiment}
    \label{fig:figure1}
\end{figure}

\begin{figure}[t]
    \centering
    \subfigure[Binary Classification]{
        \includegraphics[width=0.45\textwidth]{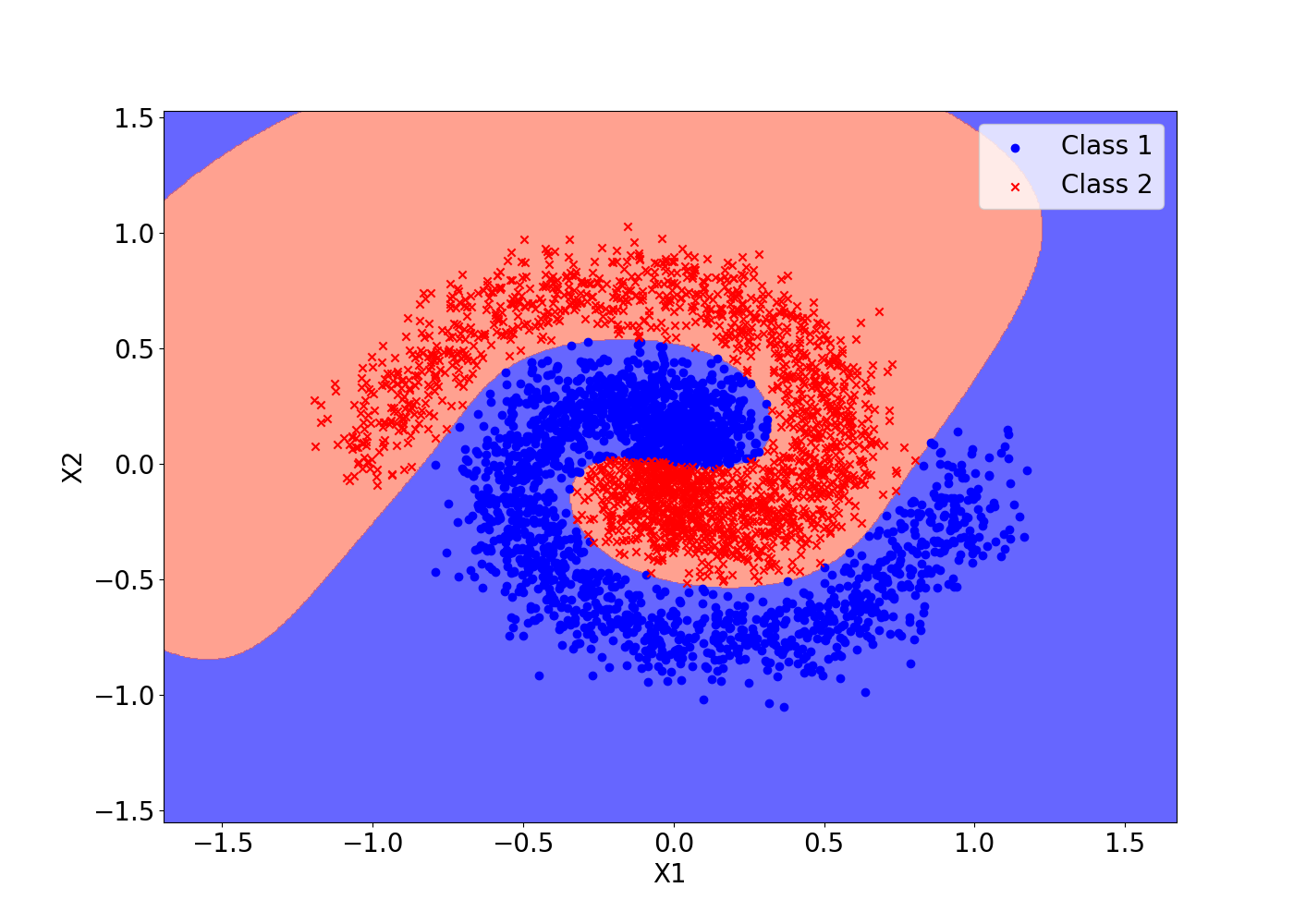}
        \label{fig:r1}
    }\hfill
     \subfigure[Proposed method (experiment \ref{item:a})]{
        \includegraphics[width=0.45\textwidth]{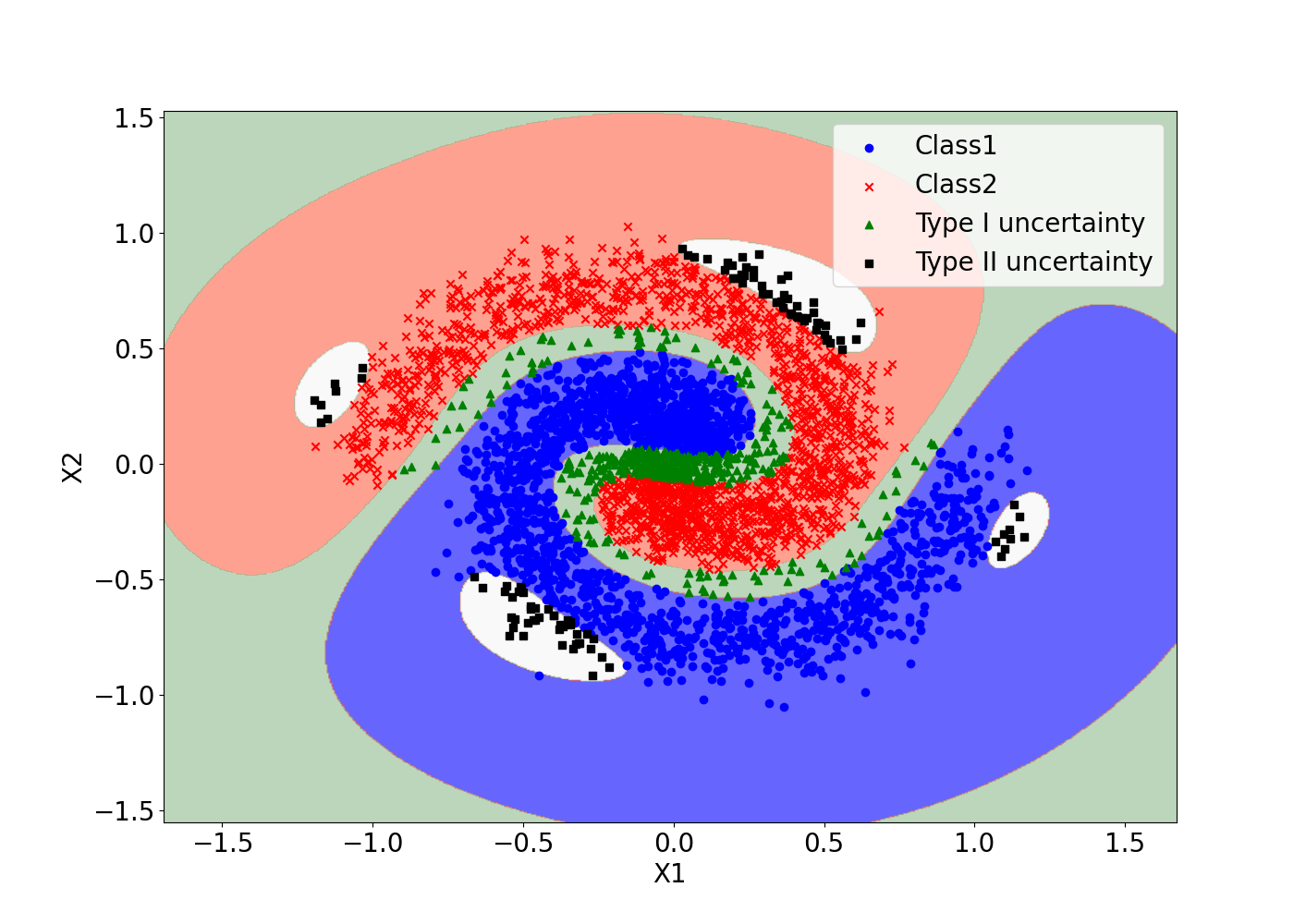}
        \label{fig:r2}
    }\\
    \subfigure[Proposed method (experiment \ref{item:b})]{
        \includegraphics[width=0.45\textwidth]{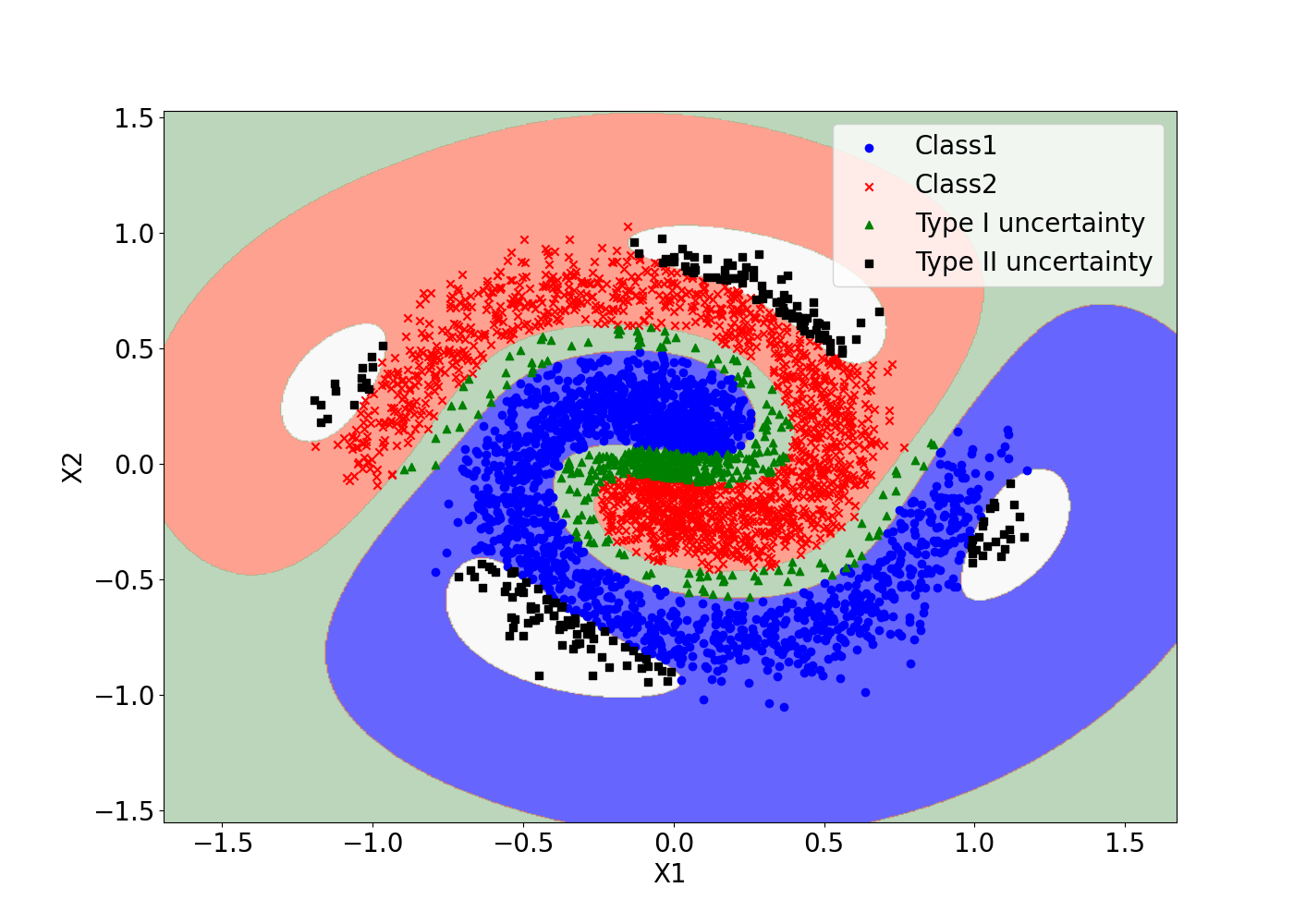}
        \label{fig:r3}
    }\hfill
    \subfigure[Proposed method (experiment \ref{item:c})]{
        \includegraphics[width=0.45\textwidth]{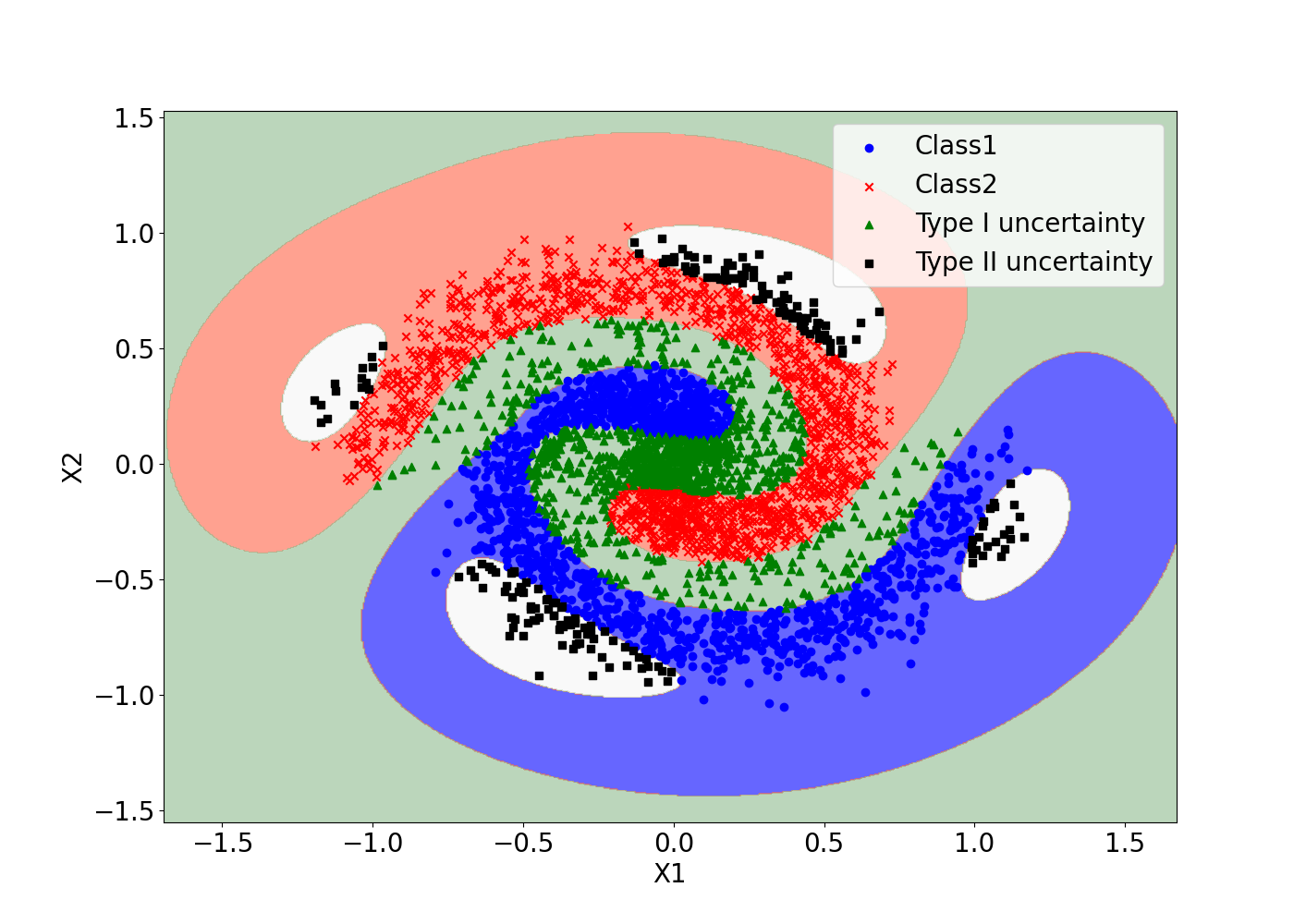}
        \label{fig:r4}
    }
    \caption{Visualization of class regions when applying the proposed method 
    and when applying conventional binary classification. Triangles represent type I uncertainty, while squares represent type II uncertainty}
    \label{fig:figure2}
\end{figure}
The proposed method is applied to the binary classification problem for spiral patterns, as illustrated in 
Figure \ref{fig:Fig1}. Class 1 is denoted as $C_1$, and Class 2 as $C_2$. The purpose of this experiment is 
to visualize class regions. Therefore, all the data in Figure \ref{fig:Fig1} are used as training data.
The SVM \citep{13,14} is 
utilized as the classifier for binary classification. The radial basis function (RBF) kernel 
$K(x, y) = \exp(-\gamma||x-y||^2)$, is employed as the kernel function of the SVM. Let $\{x_l\}^L_{l=1}$ denote 
the support vector, $\{y_l\}^L_{l=	1}$ denote the label, $\{\alpha_l \in \mathbb{R} | l=1, 2, \cdots, L\}$ 
and $b \in \mathbb{R}$ be the parameters. In this case, the discriminant function is $g(x) = \sum^L_{l=1} 
\alpha_l y_l K(x, x_l) + b$, and the value of this function is used as the test statistic.\\
\indent The histogram for each class obtained from the training data is shown in Figure \ref{fig:D1}.
Three experiments were conducted: \ref{item:a}, \ref{item:b}, and \ref{item:c}.
\vspace{5mm}
\begin{enumerate}[label=\textbf{(\roman*)}, itemsep=12pt]
\item In Test \eqref{eq1}, the acceptance region is defined as the interval between the $2.5\%$ and $97.5\%$ points, 
and in Test \eqref{eq2}, the acceptance region is defined as the interval between the $ 2.5\%$ and $97.5\%$ points.
\label{item:a}
\item In Test \eqref{eq1}, the acceptance region is defined as the interval between the $5.0\%$ and $97.5\%$ points, 
and in Test \eqref{eq2}, the acceptance region is defined as the interval between the $2.5\%$ and $95.0\%$ points.
\label{item:b}
\item In Test \eqref{eq1}, the the acceptance region is defined as the interval between the $5.0\%$ and $99.0\%$ points, 
and in Test \eqref{eq2}, the acceptance region is defined as the interval between the $1.0\%$ and $95.0\%$ points.
\label{item:c}
\end{enumerate}
\vspace{5mm}

The acceptance regions for experiments \ref{item:a}, \ref{item:b}, and \ref{item:c} are shown in Figures \ref{fig:D2}, 
\ref{fig:D3}, and \ref{fig:D4}, 
respectively, with the intervals of uncertainty indicated between the solid lines (type I uncertainty) 
and outside the dashed lines (type II uncertainty). In the context of our two types of hypothesis testing, 
if one null hypothesis is rejected and the other remains unrefuted, the data is designated as $C_1$ or 
$C_2$. To illustrate, when $H_0^1$ is accepted in test \eqref{eq1} but $H_0^2$ is rejected in test \eqref{eq2}, the data is classified as $C_1$ \\
\indent The outcomes of the binary classification are depicted in Figure \ref{fig:r1}. The results of experiments 
\ref{item:a}, \ref{item:b}, and \ref{item:c} are shown in Figures \ref{fig:r2}, \ref{fig:r3}, and \ref{fig:r4}, respectively. In the case of the 
conventional binary 
classification (Figure \ref{fig:r1}), the data near the discrimination boundary are classified as $C_1$ or $C_2$. 
Conversely, the proposed method categorizes data near the discrimination boundary as uncertain data, while 
the remaining data are classified as $C_1$ or $C_2$. A comparison of these results reveals that the proposed 
method effectively distinguishes between data while preserving the classification of data whose basis is 
ambiguous. This outcome aligns with our intuitive understanding.

\section{Applicatioin to medical image diagnosis support}\label{sec4}
In this section, we will apply the proposed method to the task of binary classification of 
pneumonia from chest X-ray images using deep learning. Many studies employ deep learning to 
perform binary classification for pneumonia \citep{15,16,17,18}. The models employed in these studies are incapable of accurately differentiating between images of individuals
who can be managed with follow-up observation and those who require additional intervention. 
Consequently, even cases that could be resolved with simple follow-up may necessitate 
unnecessary attention, potentially imposing an excessive burden on the individuals involved in 
subsequent process. Therefore, it is imperative to consider uncertainty when applying these models 
in clinical settings \citep{19,20,21,22}.

\subsection{Data preprocessing}\label{subsec4}
The dataset utilized in this study is composed of chest X-ray images (Pneumonia) obtained from Kaggle \citep{23}. The dataset 
consists of a total of 5,856 chest X-ray images, which are divided into training and test datasets. The 
training dataset contains 5,232 images, of which 3,883 are positive cases of pneumonia and 1,349 are negative 
cases. The test dataset contains 623 images, with 390 positive cases and 234 negative cases. Given the 
variation in image dimensions, the images are resized to a uniform size of $256 \times 256$ pixels. 
Subsequently, the pixel values are normalized within the range of 0 to 1.

\subsection{DenseNet}\label{subsec5}
\begin{figure}[t]
    \centering
    \subfigure[DenseNet architecture]{
        \includegraphics[width=0.45\textwidth]{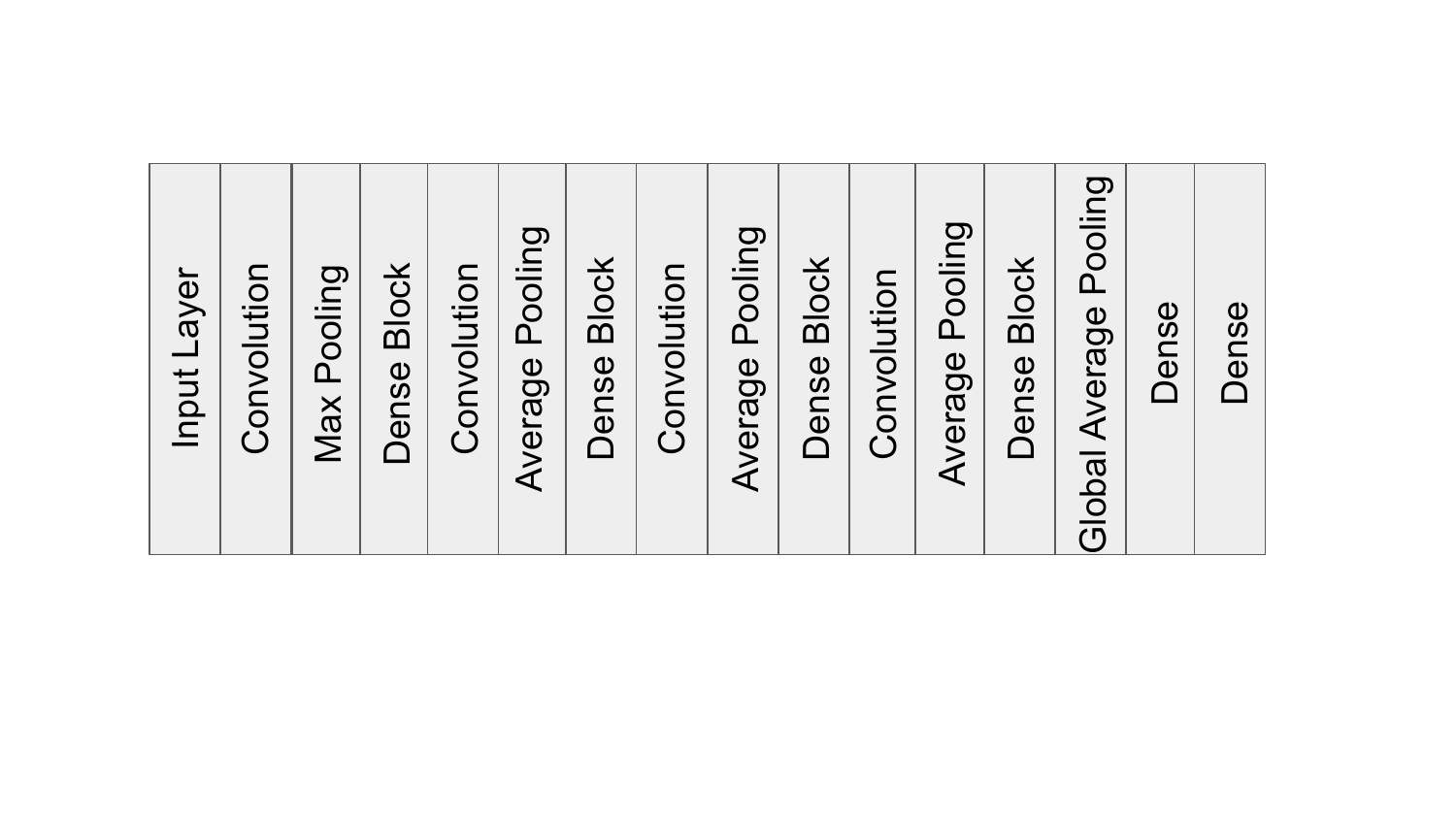}
        \label{fig:a1}
    }\hfill
    \subfigure[Dense Block]{
        \includegraphics[width=0.45\textwidth]{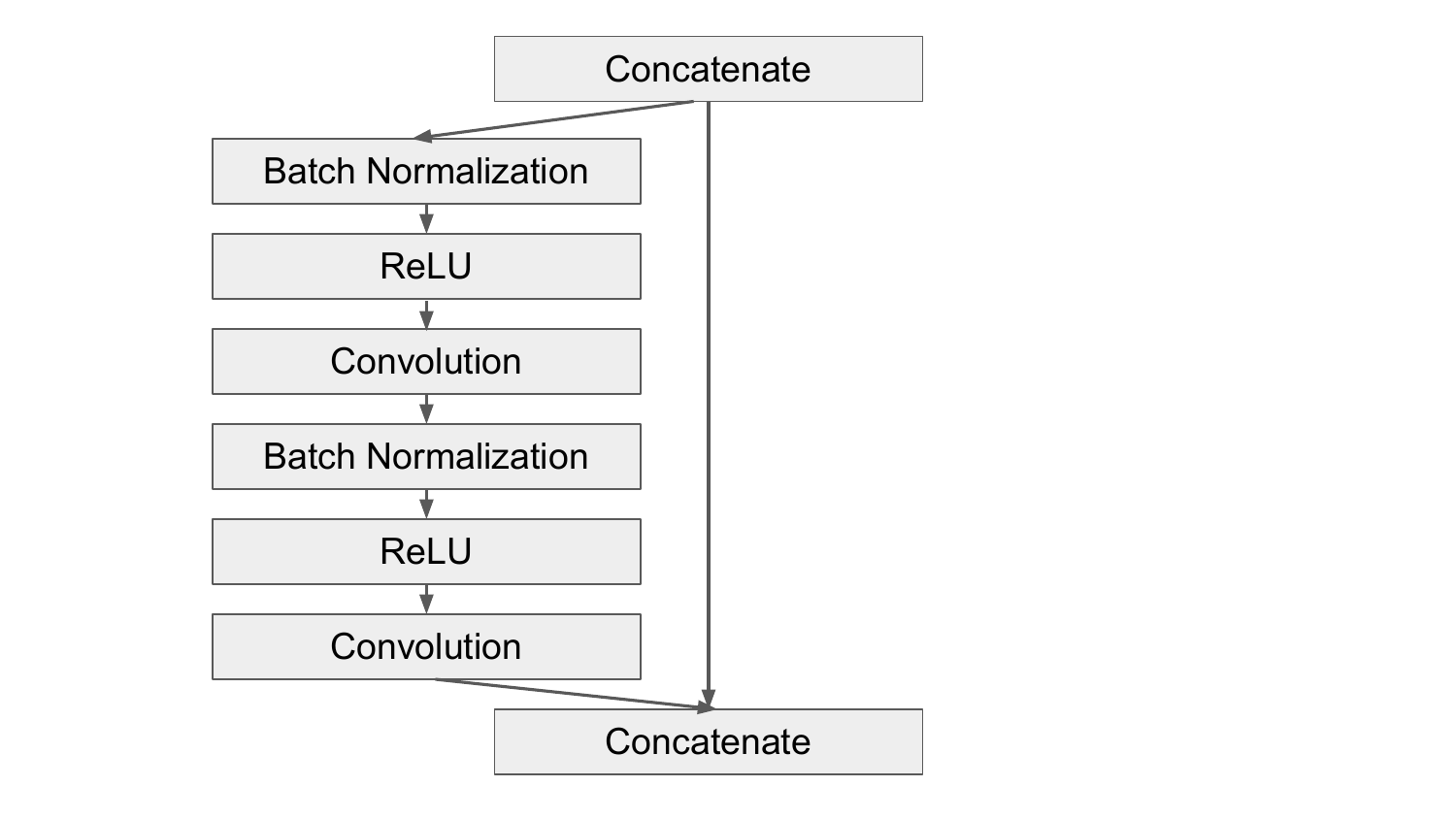}
        \label{fig:a2}
    }
    \caption{The basic architecture of DenseNet and Dense Block}
    \label{fig:figure3}
\end{figure}

DenseNet \citep{24,25,26,17} is utilized as the feature extractor. The overall structure of the network is 
illustrated in Figure \ref{fig:a1}, and certain Dense Block contained within the network are depicted in 
Figure \ref{fig:a2}. Let $z_l$ denote the output of the $l$-th layer of the Dense Block. The input to 
layer $l$ is the output values of the feature maps of all the preceding layers: 

\begin{align}
    z_l = H(\lbrack z_0,z_1,\cdots, z_{l-1} \rbrack),
\end{align}
where $\lbrack z_0, z_1, \cdots, z_{l-1}\rbrack$ is the concatenation of these feature maps across the respective layers, and $H(\cdot)$ is a composite function that acts in the order of
 Batch Normalization (BN), Rectified Linear Unit function (ReLU), and Convolution.

\subsection{Loss function}\label{subsec6}
The number of data used in this experiment is biased (see Section \ref{subsec4}). Although the training data in 
this experiment contains more positive images than negative images, no prior information about the test 
images should be assumed when employing our method in image diagnostic support. To address this issue, 
the loss function is weighted by the importance of the data. The sample sizes are denoted by $N$, and $N_1$ 
and $N_2$ for the negative and positive samples, respectively. The correct label, $y_n$, is defined as $0$ 
or $1$. The predicted probability for the label is defined as $\hat{y}_n$. The loss function with importance 
weight, which is based on the cross-entropy method commonly used in classification problems \citep{28,29}, is given as follows:
\begin{align}
    \mathcal{L} = -\sum_{n=1}^{N}w_n\lbrace y_n\log\hat{y}_n+(1-y_n)\log(1-\hat{y}_n)\rbrace, \nonumber
    \end{align}
    where
    \begin{align}
        w_n = 
        \begin{cases}
        \frac{N_2}{N}, & \text{if}\quad y_n=0\\
        \frac{N_1}{N}, & \text{if}\quad y_n=1.
        \end{cases}\nonumber
    \end{align}

\subsection{Experiment method}\label{subsec7}
As delineated in Section \ref{subsec5}, the DenseNet-121 \citep{17} is employed as the feature extractor. This network is 
pre-trained on ImageNet, and fine-tuning is achieved through the utilization of training images. 
The Adam optimization method is employed for parameter tuning. The learning rate is set to $1.0 \times 10^{-3}$, and the Adam optimizer is used. 
The test statistic, denoted by $g(x)$, satisfies the following equation: 
$\sigma(x) = 1 / (1 + \exp(-g(x)))$ (see Section \ref{subsec2}). 
The binary classification of negative or positive is performed using two types of hypothesis tests, as outlined 
in Section \ref{sec3}. However, the distinction between the types of uncertainty defined in Section \ref{sec3} is not made in 
this section. Let $\alpha$ denote the significance level. In this experiment, the parameter $\alpha$ is 
employed to delineate the two types of hypothesis testing at a common significance level. The experiment is 
conducted for values of $\alpha = 1.0\%$, $2.5\%$, and $5.0\%$. The performance of the experiment is evaluated 
using metrics such as coverage, accuracy, recall, precision, specificity, and the F1-score. Coverage is 
defined as the ratio of test data that produces results of the prediction, and is expressed as follows:
\begin{align*}
    Coverage = 1 - \frac{\#(Uncertainty\quad data)}{\#(ALL\quad data)}.
\end{align*}
The accuracy, recall, precision, specificity, and F1-score are calculated using data that was not 
judged to be uncertain.

\subsection{Result}\label{subsec8}
\begin{figure}[tb]
    \centering
    \subfigure[$\alpha=1.0\%$]{
        \includegraphics[width=0.30\textwidth]{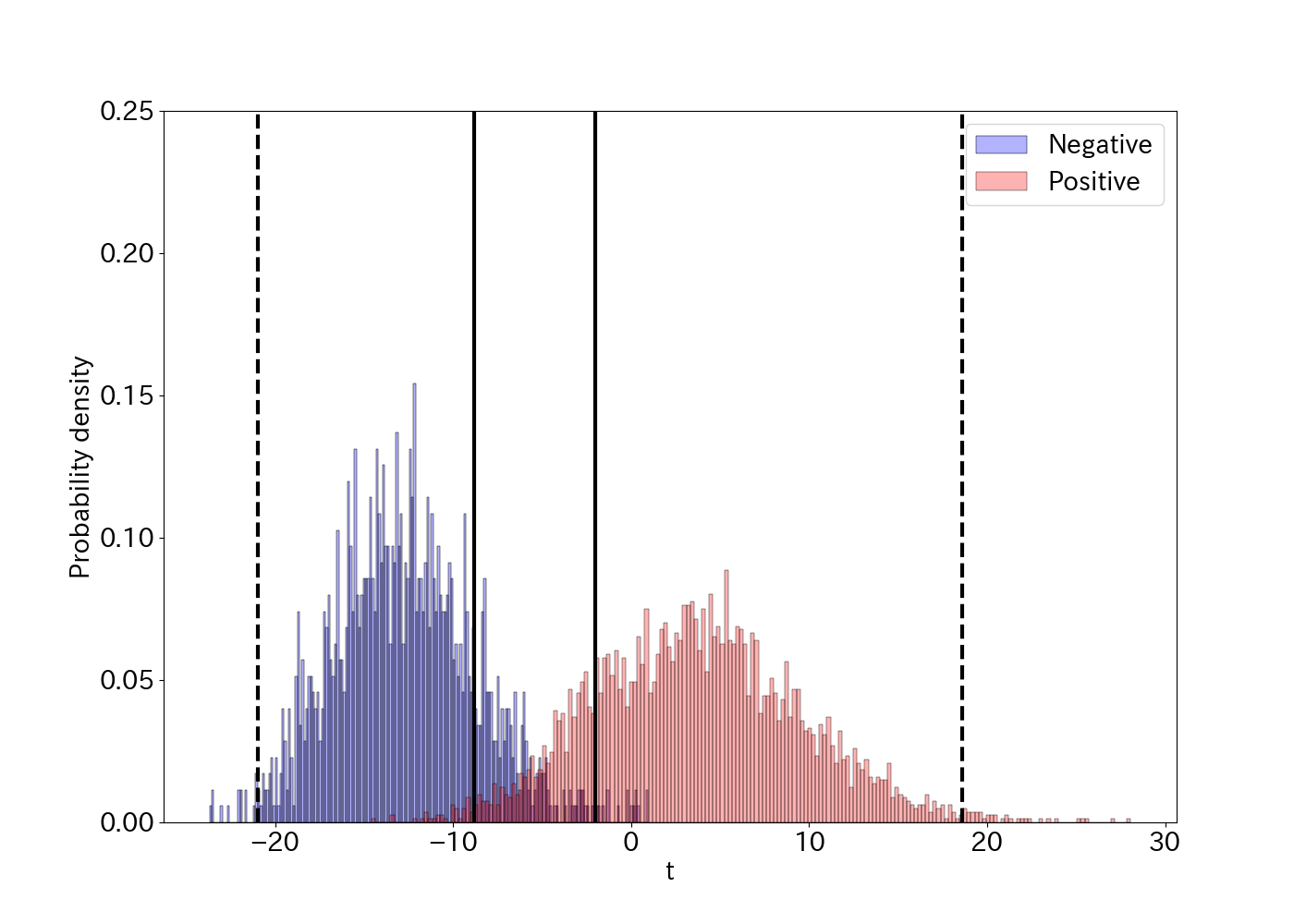}
        \label{fig:m1}
    }
    \subfigure[$\alpha=2.5\%$]{
        \includegraphics[width=0.30\textwidth]{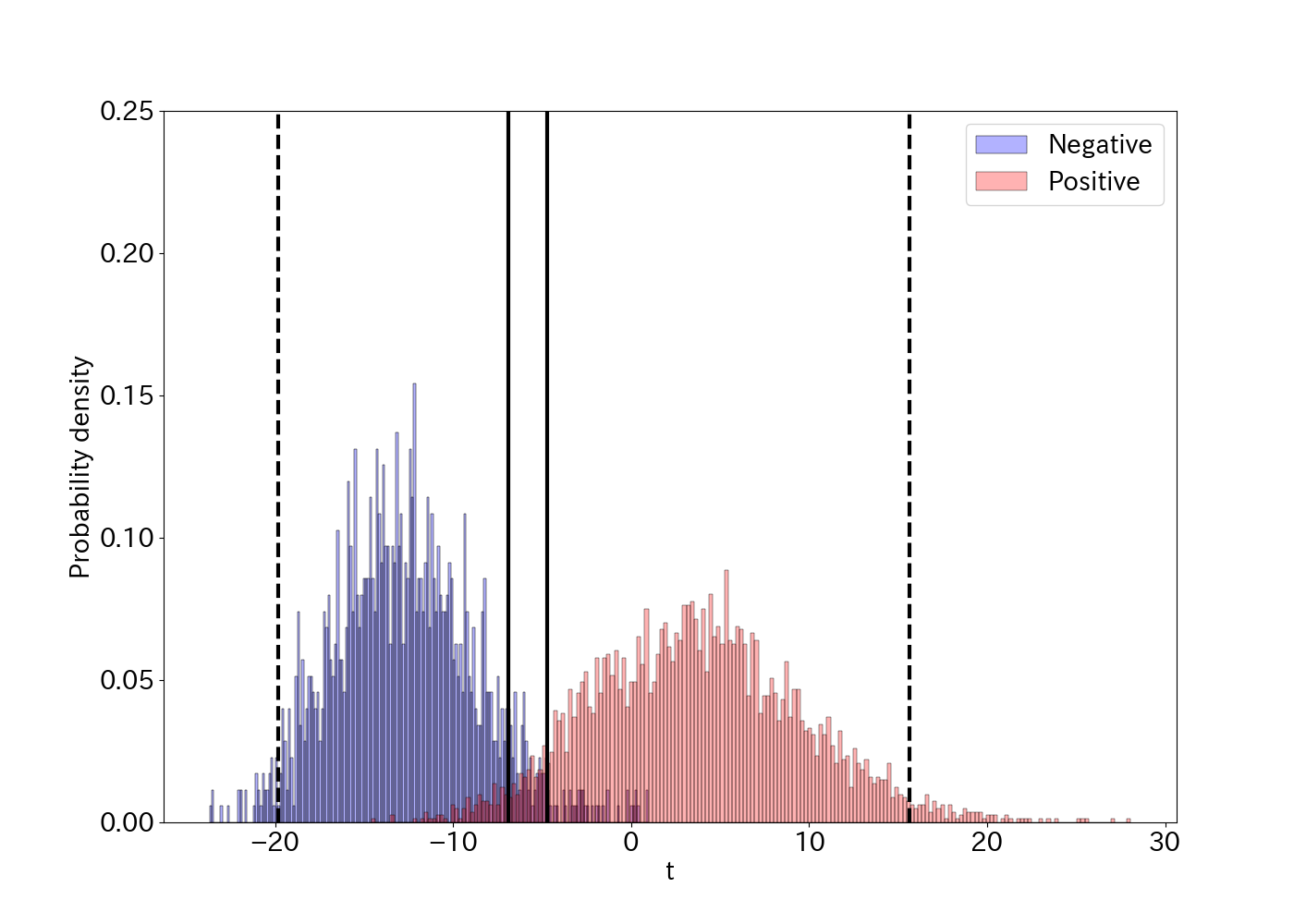}
        \label{fig:m2}
    }
    \subfigure[$\alpha=5.0\%$]{
        \includegraphics[width=0.30\textwidth]{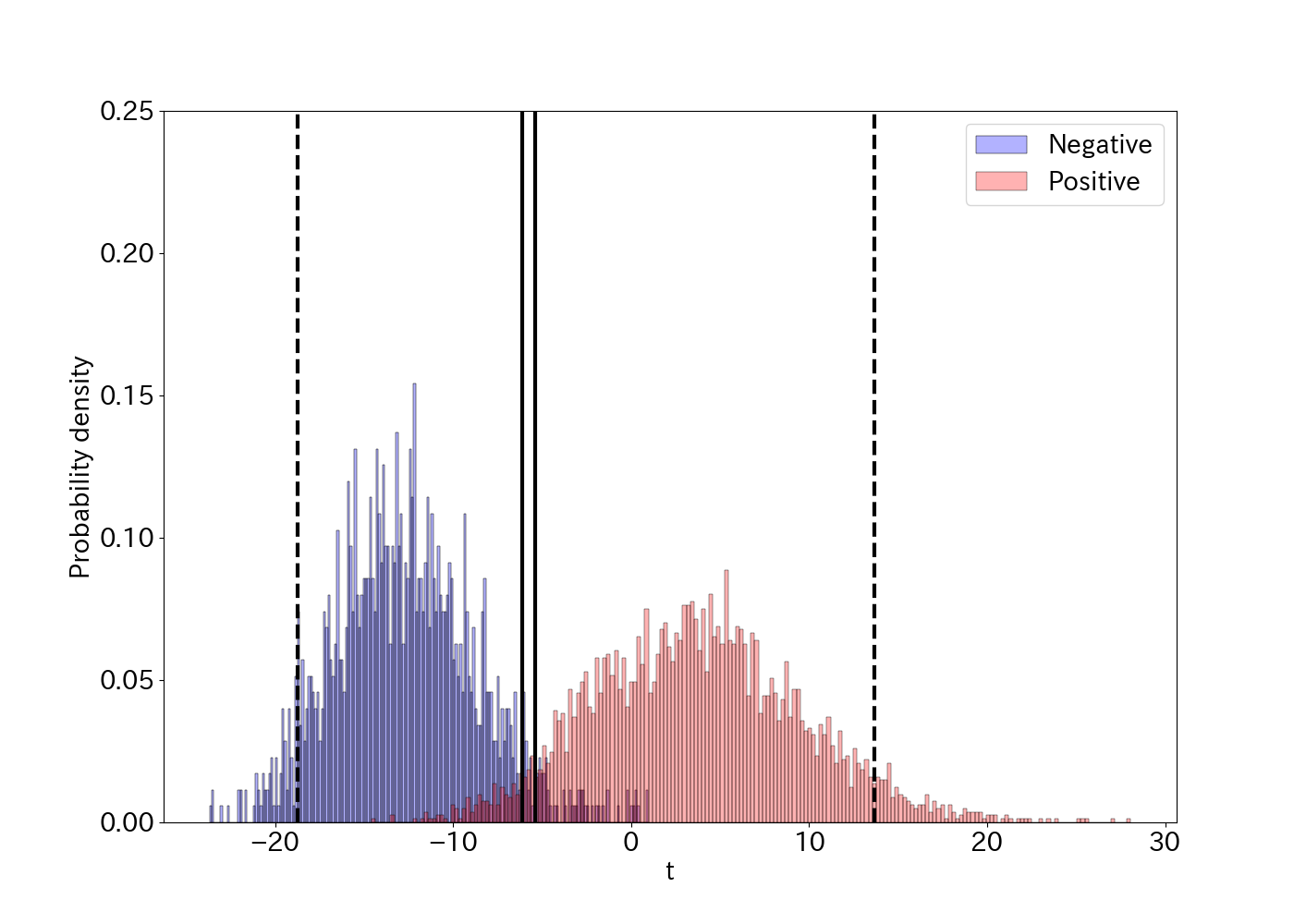}
        \label{fig:m3}
    }
    \caption{The acceptance regions of the test statistics at significance levels $\alpha=1.0\%$, $2.5\%$, and $5.0\%$. The left histogram represents the distribution of the test statistics for negative data, while the right histogram represents the distribution of the test statistics for positive data}
    \label{fig:figure4}
\end{figure}
    For the significance level $\alpha$ of $1.0\%$, $2.5\%$, and $5.0\%$, the acceptance regions of the test statistics corresponding to each level are shown in Figures \ref{fig:m1}, \ref{fig:m2}, and \ref{fig:m3}. It was observed that the range of uncertainty interval tended gradually to be narrow, as the significance level increased from $1.0\%$ to $5.0\%$. The experimental results for each significance level are shown in Table \ref{tab2}. As the significance level increases, the coverage increases, while accuracy and other evaluation metrics tend to decrease.
\begin{table}[tb]
    \caption{Experimental results at significance levels $\alpha = 1.0\%$, $2.5\%$, and $5.0\%$}\label{tab2}
    \begin{tabular*}{\textwidth}{@{\extracolsep\fill}ccccccc}
    \hline
    $\alpha$($\%$)  & coverage(\%) & Accuracy(\%) & Recall(\%) & Precision(\%) & Specificity(\%)  & F1-Score(\%) \\ \hline
    1.0  & 83.97  & 99.24   & 99.07                   & 99.68                   & 99.37 & 99.37    \\ 
    2.5 & 93.11  & 98.68   & 97.80                   & 99.44                   & 99.07 & 98.61    \\ 
    5.0  & 93.91  & 97.39   & 96.97                   & 98.87                   & 98.10      &     97.92     \\ \hline
    \end{tabular*}
\end{table}
\section{Discussion and conclusions}
\indent In this paper, we propose an approach to classification that incorporates uncertainty by introducing two types of hypothesis testing. In previous studies, the quantification of uncertainty required the use of resampled or validation data, as well as the tuning of multiple hyperparameters. In contrast, the proposed method determines thresholds using only training data, significantly reducing computational and temporal costs compared to conventional methods.
Furthermore, as described in Section \ref{sec2}, the proposed approach establishes thresholds based on the significance level defined by hypothesis testing, providing a solid theoretical foundation. Another feature of the proposed method is its ability to leverage histograms to visually capture ambiguous regions between classes while effectively detecting uncertain data.\\
\indent Two experiments were conducted in sections \ref{sec3} and \ref{sec4} of this paper. The results of these experiments confirmed that as the uncertainty interval widens, the process classifies data into one of the two classes only when the data can be judged with higher confidence. Additionally, it was revealed that the coverage does not necessarily decrease as the significance level increases.
This is because the classification thresholds are determined by the $\alpha$ quantile and $1-\alpha$ quantile corresponding to the significance level $\alpha$. Consequently, when applying the proposed method, it is necessary to adjust the significance level to an appropriate value according to the specific context.\\
\indent While this study focuses on applications in the medical field, the proposed method is expected to be applicable to other domains as well.
\bibliography{sn-bibliography}
\end{document}